\documentclass[sigconf]{acmart}

\usepackage{array}
\usepackage{stfloats}
\usepackage{url}
\usepackage{verbatim}
\usepackage{bbding}
\usepackage{graphicx}
\usepackage{textcomp}
\usepackage{xcolor}
\usepackage{multirow}
\usepackage{booktabs}
\usepackage{makecell}
\usepackage{pifont}
\usepackage{tcolorbox}
\usepackage{colortbl}
\usepackage{geometry}
\usepackage{algorithm}
\usepackage{algorithmic}
\usepackage[algo2e]{algorithm2e}
\newcommand{\eg}{\textit{e.g.}}
\newcommand{\ie}{\textit{i.e.}}
\newcommand{\our}{DeGTA\xspace}

\newsavebox\CBox

\definecolor{dark2green}{rgb}{0.1, 0.65, 0.3}
\definecolor{dark2orange}{rgb}{0.9, 0.4, 0.}
\definecolor{dark2purple}{rgb}{0.4, 0.4, 0.8}

\usepackage{dsfont}

\usepackage{hyperref}
\usepackage{tablefootnote}
\usepackage{colortbl}
\definecolor{LightCyan}{rgb}{0.88,1,1}
\definecolor{Gray}{gray}{0.92}
\definecolor{columbiablue}{rgb}{0.61, 0.87, 1.0}
\definecolor{mayablue}{rgb}{0.45, 0.76, 0.98}
\definecolor{flamingopink}{rgb}{0.99, 0.56, 0.67}

\newcommand{\yes}{\textcolor{mayablue}{\ding{51}}}

\newcommand{\no}{\textcolor{gray}{\ding{55}}}

\AtBeginDocument{%
  }

\setcopyright{acmlicensed}
\copyrightyear{2025}
\acmYear{2025}
\setcopyright{acmlicensed}\acmConference[KDD '25]{Proceedings of the 31st ACM
SIGKDD Conference on Knowledge Discovery and Data Mining V.1}{August 3--7,
2025}{Toronto, ON, Canada}
\acmBooktitle{Proceedings of the 31st ACM SIGKDD Conference on Knowledge
Discovery and Data Mining V.1 (KDD '25), August 3--7, 2025, Toronto, ON, Canada}
\acmDOI{10.1145/3690624.3709223}
\acmISBN{979-8-4007-1245-6/25/08}

\begin{document}

\title{Graph Triple Attention Network: A Decoupled Perspective}

\author{Xiaotang Wang}
\affiliation{%
  \institution{Huazhong University of Science and Technology}
  \city{Wuhan}
  \country{China}
  }
\email{m202273915@hust.edu.cn}

\author{Yun Zhu}
\authornote{Corresponding author.}
\affiliation{%
  \institution{Zhejiang University}
  \city{Hangzhou}
  \country{China}
  }
\email{zhuyun_dcd@zju.edu.cn}

\author{Haizhou Shi}
\affiliation{%
 \institution{Rutgers University}
 \city{New Brunswick}
 \state{New Jersey}
 \country{US}
 }
 \email{haizhou.shi@rutgers.edu}

\author{Yongchao Liu}
\affiliation{%
  \institution{Ant Group}
  \city{Hangzhou}
  \country{China}
}
\email{yongchao.ly@antgroup.com}

\author{Chuntao Hong}
\affiliation{%
  \institution{Ant Group}
  \city{Hangzhou}
  \country{China}
}
\email{chuntao.hct@antgroup.com}

\begin{abstract}
Graph Transformers (GTs) have recently achieved significant success in the graph domain by effectively capturing both long-range dependencies and graph inductive biases. However, these methods face two primary challenges: (1) \emph{multi-view chaos}, which results from coupling multi-view information (positional, structural, attribute), thereby impeding flexible usage and the interpretability of the propagation process. (2) \emph{local-global chaos}, which arises from coupling local message passing with global attention, leading to issues of overfitting and over-globalizing.
To address these challenges, we propose a high-level decoupled perspective of GTs, breaking them down into three components and two interaction levels: positional attention, structural attention, and attribute attention, alongside local and global interaction. Based on this decoupled perspective, we design a decoupled graph triple attention network named \our, which separately computes multi-view attentions and adaptively integrates multi-view local and global information. This approach offers three key advantages: enhanced interpretability, flexible design, and adaptive integration of local and global information.
Through extensive experiments, \our achieves state-of-the-art performance across various datasets and tasks, including node classification and graph classification. Comprehensive ablation studies demonstrate that decoupling is essential for improving performance and enhancing interpretability.

\end{abstract}



\maketitle
\section{Introduction}


Recently, Graph Transformers (GT)~\citep{gt,egt,graphgps} have shown great potential in handling graph-structured data such as social networks~\citep{socialnetwork}, drug discovery~\citep{drug}, and traffic networks~\citep{traffic}. The success of GTs is built upon their ability to leverage inductive bias and handle long-range dependencies simultaneously. For example, Graphormer~\citep{graphormer} utilizes global attention to capture long-range dependencies and uses structural biases to influence attention scores, thereby introducing inductive bias. GraphGPS~\citep{graphgps} employs GNNs before global attention to enhance graph topology information. Besides, these methods will compensate positional encodings and structural encodings into node features to enhance the performance.
However, existing GTs mainly face two challenges while integrating diverse information:

1) \emph{multi-view chaos}: Multi-view encodings, such as positional encodings, structural encodings, and attribute encodings, are crucial for building a GT~\citep{gt}. Existing methods couple these types of information to obtain node features, effectively harnessing graph topology and inductive biases. However, multi-view chaos constrains the separability of positional, structural, and attribute information during propagation, thus impeding flexible usage and the interpretability of the propagation process.

2) \emph{local-global chaos}: Existing GTs couple local message passing and global attention, which complicates the adjustment of importance weights between local and global information. This uniformity potentially precipitates issues of overfitting and over-globalizing~\citep{overglobalizing}. In some graphs, local information may be crucial, while in others, global information may dominate. Coupling these sources of information makes it challenging to adapt the weights appropriately and lacks interpretability of which information contributes to success.

To address the aforementioned challenges, we propose a high-level decoupled perspective of Graph Transformers, breaking down GTs into three components and two message interaction levels, namely: (i) \textbf{Positional Attention}: Self-attention on positional information. (ii) \textbf{Structural Attention}: Self-attention on structural information. (iii) \textbf{Attribute Attention}: Self-attention on attribute information.
Additionally, the interaction levels are:
(a) \textbf{Local-level}: Aggregating messages between neighbors.
(b) \textbf{Global-level}: Aggregating messages between all nodes.
Based on this decoupled perspective, we design a decoupled graph triple attention network, coined as \our, that separately computes multi-view attentions, including positional attention, structural attention, and attribute attention, and adaptively integrates local and global information. This approach offers three key advantages:

1. \emph{Enhanced Interpretability}: Aggregation becomes more interpretable as we can visualize the attention scores separately, enabling analysis of which information contributes the most.

2. \emph{Flexible and Adaptive Design}: The attention mechanism for multi-view information can be flexibly designed and combined in an adaptive manner.

3. \emph{Adaptive Integration of Local and Global Information}: By employing a global sampling strategy, \our can capture long-range dependencies with global positional and structural attention, and then adaptively integrate global and local information.

Our contributions can be summarized as follows:
\begin{itemize}
    \item We propose a decoupled perspective for attention in graphs. From this perspective we identify two main challenges of information coupling in GTs and provide a framework for designing novel decoupled models.
    \item Based on this decoupled perspective, we introduce \our, a decoupled graph triple attention network that enhances interpretability, enables flexible design of multi-view attention, and adaptively integrates local and global information.
    \item Through extensive experiments, \our achieves state-of-the-art performance across various datasets and tasks, such as node-level classification and graph classification. Comprehensive ablation studies demonstrate that decoupling is essential for improving performance and enhancing interpretability.
\end{itemize}

\section{Related Work}

\subsection{Graph Attention Networks \& GTs}

Graph attention-based networks endeavor to ascertain the relational significance between node pairs. These models can be classified into two main categories: (1) edge-attention, exemplified by the Graph Attention Network (GAT)~\cite{gat} and its variants~\cite{cat,gatv2}, in which each source node aggregates features from its neighbors based on the deduced importance of the edges; and (2) hop-attention~\citep{dagnn,gprgnn}, which discerns the relative importance of each hop. Hop-attention models apply attention weights to each hop's information and then compute node features through a weighted summation of different hop information.


\sloppy
The above models conduct message passing based on graph topology, \eg, MPNN-based~\citep{mpnn,gcn,rosa,unigap}, which can be considered a local interaction between nodes. It's well known that MPNN-based GNNs face challenges of over-smoothing~\citep{oversmoothing}, oversquashing~\citep{oversquashing}, and limited expressive power limitations~\citep{expressivepower}. To extend the receptive field, Graph Transformers (GTs)~\citep{gt,egt} have emerged, utilizing global attention combined with positional and structural information to simultaneously capture long-range dependencies and graph inductive bias. 
For instance, Graphormer~\cite{graphormer} uses centrality encoding to enhance structural information and spatial matrix as positional biases, SAN~\cite{san} uses a learnable Laplacian PE as input to a branch Transformer layer, and GRIT~\cite{grit} uses a random walk encoding to integrate positional information. 
GraphGPS~\cite{graphgps} explicitly integrate various other types of MPNN modules into their architectures. Other works such as NAGphormer~\cite{nagphormer} and SAT~\cite{sat} use subgraph to introduce neighborhood inductive biases to node features. However, GTs face two significant problems while integrating different categories of information: (1) local-global chaos, and (2) multi-view chaos. These issues make it challenging for GTs to adapt weights appropriately, lack interpretability regarding which information contributes to success, and impede flexible usage and the interpretability of the propagation process.

In this work, we propose a high-level decoupled perspective of GTs. Based on this perspective, we design a novel decoupled graph triple attention network to address the above problems.

\subsection{PE and SE}
In the graph domain, PE and SE have been studied for enhancing the expressive power and performance of both MPNNs and GTs, particularly for GTs to introduce inductive bias of graph. For instance, some works incorporate Laplacian encoding~\citep{san} random walk encoding~\citep{grit}, shortest-path-distance~\citep{distanceencoding} or centrality encoding~\citep{graphormer} as PE/SE to capture important positional or structural relationships between nodes, with certain studies emphasizing the learnability of PE/SE. However, there is a paucity of work that clearly distinguishes between the definitions of PE and SE, many so-called positional or structural encodings comprising both positional and structural information, resulting in the coupling of information. Additionally, these encodings are often concatenated or integrated with attribute encodings to compute overall attention, leading to multi-view chaos. Section~\ref{subsec:revisitworks} contains a detailed revisit including the usage of PE and SE in previous works.

\section{Revisiting Graph Attention Mechanism: A Decoupled Perspective}\label{sec:decoupled}
In this section, we introduce our decoupled perspective for the graph attention mechanism from two core aspects: multi-view attentions and message interactions.
Then, we revisit previous work from the decoupled perspective and draw conclusions.

\begin{table*}[]
\caption{Revisiting previous work through our decoupled perspective. If the last two columns of "Decoupling" are not checked, the ticked attentions are considered coupled to one another.}
\vspace{-1em}
\label{tab:decouple}
\centering
\scalebox{0.9}{
{\tabcolsep0.1in
\begin{tabular}{lcccccccccccc}
\toprule[1.2pt]
Attention Type          && \multicolumn{2}{c}{Positional Attention}&& \multicolumn{2}{c}{Structural Attention} && \multicolumn{2}{c}{Attribute Attention}  && \multicolumn{2}{c}{Decoupling}     \\
\cline{1-1} \cline{3-4} \cline{6-7} \cline{9-10} \cline{12-13}
interaction          && local                 & global            & & local               & global              & & local               & global      && multi-view    & local-global                       \\ 
\midrule
GCN~\citep{gcn}       && \no                 & \no                 & & \no                 & \no                 & & \yes                & \no         && \no  & \no                     \\
GAT~\citep{gat}       && \no                 & \no                 & & \no                 & \no                 & & \yes                & \no         && \no  & \no                     \\
ADSF~\citep{adsf}      && \no                 & \no                 & & \yes                & \no                 & & \yes                & \no         && \yes & \no                     \\
GT~\citep{gt}        && \no                 & \yes                & & \no                 & \no                 & & \no                 & \yes        && \no  & \no                     \\
Graphormer~\citep{graphormer}&& \no                 & \yes                & & \no                 & \yes                & & \no                 & \yes        && \no  & \no                     \\
SAN~\citep{san}       && \no                 & \yes                & & \no                 & \no                 & & \no                 & \yes        && \yes & \no                     \\
LSPE~\citep{lspe}      && \yes                & \no                 & & \no                 & \no                 & & \yes                & \no         && \yes & \no                     \\
GraphGPS~\citep{graphgps}  && \no                 & \yes                & & \no                  & \yes               & & \yes                & \yes        && \no  & \yes                    \\
SAT~\citep{sat}       && \no                 & \yes                & &  \no                & \yes                & & \no                 & \yes        && \no  & \no                     \\
NAGphormer~\citep{nagphormer} && \yes                & \no                 & &  \yes                & \no                & & \yes                & \no         && \no  & \no                     \\ 
\midrule
\our      && \yes                & \yes                & & \yes                & \yes                & & \yes                & \yes        && \yes & \yes                    \\
\bottomrule[1.2pt]
\end{tabular}
}}
\vspace{-1em}
\end{table*}

\subsection{Decoupled Perspective}
\subsubsection{Decoupled Perspective of Multi-View Attentions}\label{subsec:attn}
In this work, we refer to positional, structural, and attribute information as multi-view information. First, we provide definitions of these types of information, as few previous works~\cite{graphgps} offer clear concepts of them. Subsequently, we introduce several optional encodings from our decoupled perspective. 

(i) Structural Encoding (SE) focuses on a node's capacity to perceive its surrounding structure. A node does not concern itself with the specific identities of its neighboring nodes; rather, it focuses exclusively on topological information, such as degree information, the shape of its subgraph, and other topological characteristics like triangle counting and cycle counting. 
In our method, we provide several strategies to achieve this objective: (1) Random-Walk Structural Encoding (RWSE)~\cite{lspe}, (2) Node Degree Encoding (DSE)~\cite{graphormer}, and (3) Topology Counting Encoding (TCSE)~\cite{tcse}. The details of these encodings can be found in Appendix~\ref{sec:appendidxb}.

(ii) Positional Encoding (PE) focuses on the capability of nodes to perceive their relative positions with respect to other specific nodes and their positions within the entire graph. In contrast to structural information, positional information is more specific, as it can discern the identity information of other nodes. For instance, it can sense the distance to other nodes and intersections with other nodes within n-hop subgraphs. 
We provide several off-the-shelf strategies that can reflect a node's distance or intersection relationships with other specific nodes: (1) Laplacian Positional Encoding (LapPE)~\cite{san,graphcontrol}, (2) Relative Random Walk Probabilities (RWPE)~\cite{grit}, and (3) Jaccard Encoding (JaccardPE)~\cite{adsf}. The introduction of these strategies can be found in Appendix~\ref{sec:appendidxb}.

(iii) Attribute Encoding (AE) aim to transform raw node attributes, such as text or images, into numeric vectors. Recently, with the significant success of LLMs in language understanding, text-attributed graphs (TAGs) have become a prominent topic in the graph domain. For TAGs, LLMs can be employed to enhance attribute encodings~\cite{llmgraph,engine}.  Thus we can select AE from original features~\citep{pyg} or LLM-enhanced features~\citep{engine,graphclip}.

Previous methods like GraphGPS~\cite{graphgps} and Graphormer~\cite{graphormer}, couple these information types to calculate attention, leading to \emph{multi-view chaos}, which impedes flexible usage and the interpretability of the propagation process. 
The clear definitions can guide us in checking if the information is adequate and help address the \emph{multi-view chaos} of attention in graphs by separately computing structural, positional, and attribute attention through different encodings of node pairs. 

\subsubsection{Decoupled Perspective of Message Interaction}\label{subsec:inter}
Message interaction means how does messages of nodes interact with each other. There are definitions of different interactions: 

(a) Local Interaction~(LI) is the information aggregation based on the original graph topology, where message passing occurs only along the edges present in the original graph.

(b) Global Interaction~(GI) refers to the message passing between all pairs of nodes whether there have edges in the original graph or not.

According to these definitions, we can identify \emph{local-global chaos} in current GTs, which makes it challenging to adapt the weights appropriately and lacks interpretability regarding which information contributes to success. Decoupling these interactions is essential for building a robust and interpretable graph model.

\subsection{Revisiting Previous Work}
\label{subsec:revisitworks}
In this subsection, we will revisit previous classical studies from our decoupled perspective. Table~\ref{tab:decouple} shows the comparisons of different methods from this viewpoint.
GCN uses AE for message passing along the graph structure and averages the aggregated neighbors. While GAT uses an adaptive attention aggregation. ADSF uses a Structural Fingerprint, indeed a Jaccard encoding as PE to compute a fixed positional attention, and adaptively integrate it with attribute attention. GT replaces GAT's local attention of AE with global attention. Graphormer uses centrality encoding as SE, and it uses spatial matrix directly as positional attentions to serve as a bias term in the self-attention module, which coupled AE and PE. SAN generates a learnable PE from a branch Transformer layer, then it concats this PE with AE, and computes global attention. This attention score is equal to the sum of positional attention and attribute attention. LSPE proposes learnable positional and structural encoding, however, its so-called structural encoding is node features and then concat with RWSE, which is equivalent to the coupling of AE and SE from our perspective, while its positional encoding is RWSE, indeed a SE as we defined. GPS explicitly splits positional encodings and structural encodings and it uses MPNN before transformer to incorporate local information into global. SAT uses subgraph extractor to couple SE and PE with AE through the subgraph information, and then input the coupled encoding to Transformer layers to get global attention. NAGphormer uses the multi-hop representation based on subgraph as the token sequence of the Transformer layer, which couples the PE, SE, and AE, and computes attention between these local representations.

As shown in Table~\ref{tab:decouple} and the above discussions, previous works struggle to decouple multi-view information and different-level message interactions, leading to \emph{multi-view chaos} and \emph{local-global chaos}. Additionally, some methods do not utilize sufficient information and interactions, resulting in sub-optimal performance. 
In order to solve these issues, we need to answer the following research questions:
\begin{tcolorbox}
\textbf{RQ1}: How to rationally utilize all the information in the graph, while avoiding the multi-view chaos?
\end{tcolorbox}

\begin{tcolorbox}
\textbf{RQ2}: How to avoid the local-global chaos while capturing the long-range dependencies and maintaining the inductive biases?
\end{tcolorbox}

We will design our model, named \our, by addressing these problems to resolve existing issues. The details are illustrated in the next section.

\section{Decoupled Graph Triple Attention Network}
\begin{figure*}[t] 
\centering 
\includegraphics[width=1\linewidth]{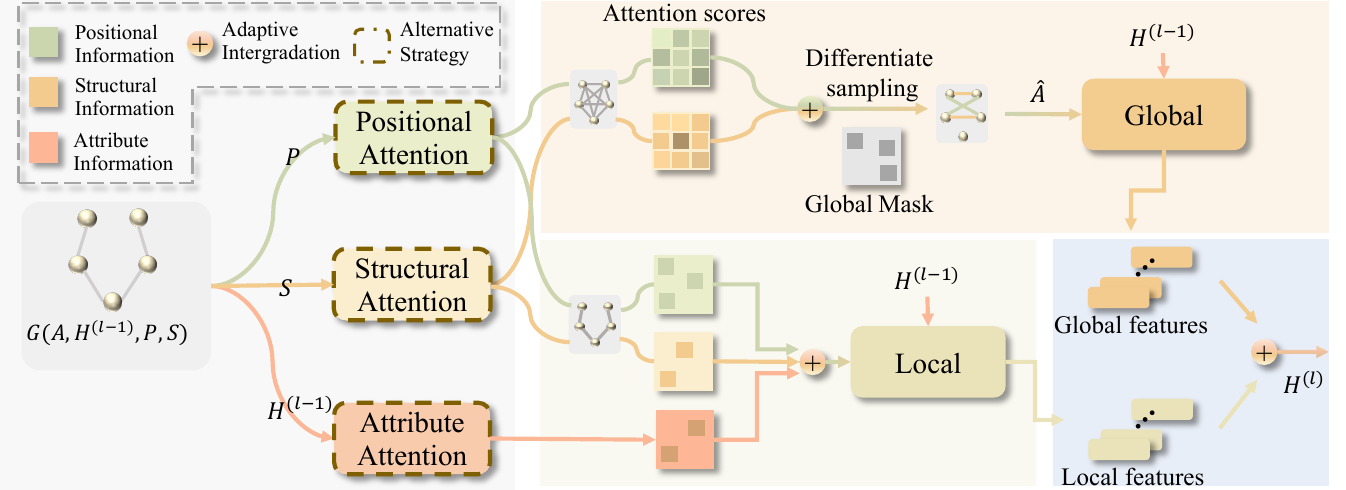} 
\vspace{-1em}
\caption{Framework of \our. The framework comprises four main components: Decoupled Multi-View Encoder, Local Channel, Global Channel, and Local-Global Integration. The strategies of encoders for multi-view and attention mechanism for local and global are all optional.}
\label{framework} 
\vspace{-1em}
\end{figure*}

In this section, we will introduce our designed model, named \our, by addressing RQ1 in Section~\ref{sec:rq1} and RQ2 in Section~\ref{sec:rq2}. Then, we discuss \our from a theoretical perspective in Section~\ref{sec:discuss}. The overall framework of our method is shown in Figure~\ref{framework}. 

\subsection{Notations}
Let $G=(\mathcal{V},{A},{X})$ denote a graph, where \( \mathcal{V} \) denote the sets of nodes, respectively. \( {X} \in \mathbb{R}^{N \times d} \) represents the node attribute feature matrix with \( N \) nodes and \( d \) attributes. \( {A} \in \mathbb{R}^{N \times N} \) is the adjacency matrix. For positional and structural information, we denote the initial PE as \( P \in \mathbb{R}^{N \times K} \), and the initial SE as \( S \in \mathbb{R}^{N \times K} \).
Additionally, we denote the diagonal degree matrix as \( D \in \mathbb{R}^{N \times N} \). The matrices with self-loops are denoted as \( \tilde{A} \in \mathbb{R}^{N \times N} \) and \( \tilde{D} \in \mathbb{R}^{N \times N} \), respectively. Thus, the graph Laplacian with self-loops is \( \tilde{L} = \tilde{D} - \tilde{A} \), the normalized adjacency matrix is \( \hat{A} = \tilde{D}^{-1} \tilde{A} \), and the symmetrically normalized graph Laplacian is \( \hat{L} = \tilde{D}^{-\frac{1}{2}} \tilde{L} \tilde{D}^{-\frac{1}{2}} \). 
For clarity in notations, we will use \( A \) to denote the original graph (local level). The sampling graph in global level will be represented by \( \hat{A} \).

\subsection{Decoupling Multi-view Attention (RQ1)}\label{sec:rq1}
To rationally utilize all the information in the graph and avoid multi-view chaos, we meticulously design multi-view encodings and decouple multi-view attentions. First, we introduce the initialization strategies for structural, positional, and attribute information. Then, we illustrate how to achieve decoupled multi-view attention.

\subsubsection{Multi-view Information Encodings}
In this subsection, we employ the available encoding strategies for positional, structural, and attribute information that introduced in Section~\ref{subsec:attn}. Because our method decouples multi-view attention, it allows for flexible selection of encoding strategies.

\paragraph{Strategy Selection.} 
Different graphs may require different positional and structural encoding strategies, making it challenging to choose a specific encoding strategy that performs best across all datasets. In this work, considering simplicity and effectiveness, we choose Jaccard Positional Encoding (JaccardPE) and Random-Walk Structural Encoding (RWSE) as our positional and structural encodings, respectively. For node attribute encodings, we use the provided node encodings by PyG~\cite{pyg}.
Exploration of more complex strategy selection will be left for future work.

\subsubsection{Decoupled Multi-view Attention.} 
To avoid multi-view chaos, unlike previous works that operate overall attention with mixed encodings, we handle them separately with independent encoders. This decoupled approach allows us to flexibly choose the type of encoders for each type of information as follows:
\begin{equation}
S^{(l)}=\mathcal{E}_{s}^{(l)}(S), \quad
P^{(l)}=\mathcal{E}_{p}^{(l)}(P), \quad
H^{(l)}=\mathcal{E}_{a}(H^{(l-1)})
\end{equation}
where $\mathcal{E}_{s} (\cdot): \mathbb{R}^{N \times K} \mapsto \mathbb{R}^{N \times d_{s}}$, $\mathcal{E}_{p} (\cdot): \mathbb{R}^{N \times K} \mapsto \mathbb{R}^{N \times d_{p}}$, and $\mathcal{E}_{a}(\cdot): \mathbb{R}^{N \times d} \mapsto \mathbb{R}^{N \times d}$ are alternative encoders for processing structural, positional and attribute encodings respectively. Note that $d_{s},d_{p} \ll d$ because $K$ is always small to control the receptive field, generally no more than 8. To retain the pure structural and positional information, we feed the initial SE and PE into each layer. For AE, to obtain high-level features, we use the output of the previous layer as input.

\begin{figure}[t] 
\centering 
\includegraphics[width=1\linewidth]{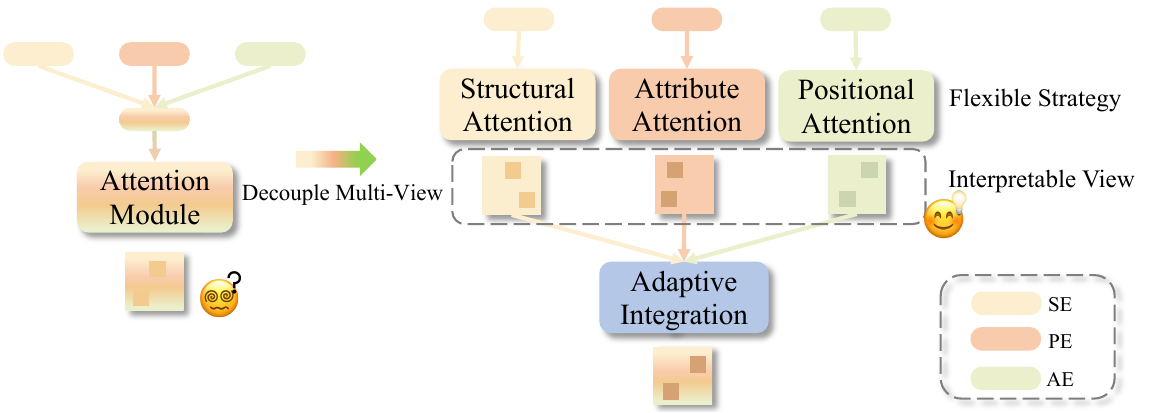}
\vspace{-2em}
\caption{Comparison of traditional attention and decoupled multi-view attention. Our method enables the flexible design of distinct attention mechanisms for various encodings, and enhances interpretability by the capacity to visualize the attention scores independently.}
\label{multiview} 
\vspace{-1em}
\end{figure}

In this work, we design a graph triple attention network that utilizes independent attention modules to process positional, structural, and attribute encodings. As shown in Figure~\ref{multiview}, this approach allows us to visualize the attention scores separately, facilitating an analysis of which information is similar in a node pair, thereby improving interpretability. We then adaptively learn the weights of the three types of attention for information aggregation, enabling an analysis of which information contributes the most through the learned weights.

\subsection{Decoupling Message Interaction (RQ2)}\label{sec:rq2}

To avoid the local-global chaos while capturing long-range dependencies and maintaining graph inductive bias (RQ2), we decouple the local and global interaction in this work by designing separate attention mechanisms for each level of interaction. This decoupling allows us to effectively manage the unique characteristics and importance of local and global information, enhancing both performance and interpretability.

\subsubsection{Decoupling Local-level Interaction.}
The local-level interaction focuses on message passing between neighboring nodes, leveraging the immediate graph topology. This interaction is crucial for capturing fine-grained local structures and details within the graph. In our work, we use MPNN-based attention methods to compute local attention as follows:
\begin{equation}
\begin{aligned}
    s_{i,j} &= q_{s}^{T}\sigma([W_{\text{str}}S_{i}\parallel W_{\text{str}}S_{j}])\\
    p_{i,j} &= q_{p}^{T}\sigma([W_{\text{pos}}P_{i}\parallel W_{\text{pos}}P_{j}])\\
    a_{i,j} &= q_{a}^{T}\sigma([W_{\text{atr}}H_{i}\parallel W_{\text{atr}}H_{j}])\\
\end{aligned}
\end{equation}
where $\cdot^T$ represents transposition and $\parallel$ is the concatenation operation, $\sigma$ is a non-linear activation, \eg, LeakyReLU. $W_{\text{str}} \in \mathbb{R}^{d' \times d_{s}}$, $W_{\text{pos}} \in \mathbb{R}^{d' \times d_{p}}$, $W_{\text{atr}} \in \mathbb{R}^{d'' \times d}$ denote feature transformation matrices, and $q_{s},q_{p} \in \mathbb{R}^{2d'}$ and $q_{a} \in \mathbb{R}^{2d''}$ are learnable weight vectors. The decoupled multi-view attention scores, \ie, $s_{i,j}$, $p_{i,j}$, and $a_{i,j}$, can assist in the interpretability through visualizing the attention scores separately.

\paragraph{Local Integration} Then, we adaptively combine the scores of local structural, positional, and attribute attention:
\begin{equation}
    z_{i,j} = \alpha p_{i,j}+ \beta s_{i,j} + \gamma a_{i,j}
\end{equation}
where \(\alpha\), \(\beta\), and \(\gamma\) are learnable weights for the multi-view attention, enabling adaptive use of decoupled multi-view information to compute the total weight of local aggregation. This adaptive mechanism allows for an analysis of which type of information contributes the most to the graph through the learned weights, enhancing both the flexibility and interpretability of the model.

Last, we normalize the attention scores and aggregate the local information through AE along the original graph:
\begin{equation}
    \hat{H}_{i}^{\text{local}}= \sum_{j\in \mathcal{N}_{i}}\hat{z}_{i,j}H_{j},\;\text{where}\;\hat{z}_{i,j} = \frac{e^{z_{i,j}}}{\sum_{k \in \mathcal{N}_{i}}e^{z_{i,k}}}
\end{equation}
where $H_{i}$, $H_{j}$ is attribute encoding of node $i$ and $j$. and $\mathcal{N}_{i}$ is the set of neighbours of node $i$.

\subsubsection{Decoupling Global-level Interaction.}
The global-level interaction captures long-range dependencies and overall graph structures by employing global attention mechanisms that can link distant nodes and aggregate information across the entire graph. To achieve this, we will sample a new graph topology that can effectively capture long-range dependencies from a global viewpoint. This new topology will enable the model to consider connections beyond immediate neighbors, ensuring comprehensive integration of global information.
First, we compute global attention by Transformer attention layer:
\begin{equation}
\begin{aligned}
    U_{s}&=\text{Softmax}(\frac{Q_{s}K_{s}^{T}}{\sqrt{d_{s}}}),\quad U_{p}=\text{Softmax}(\frac{Q_{p}K_{p}^{T}}{\sqrt{d_{p}}})
\end{aligned}
\end{equation}
where $Q_s=SW_s^q,Q_p=PW_p^q$ are the global query features of structural and positional encodings, and $K_s=SW_s^k, K_p=PW_p^k$ are the global key features of structural and positional encodings, $W_s^q,W_s^k \in \mathbb{R}^{d_{s} \times d'}$ and $W_p^q,W_p^k \in \mathbb{R}^{d_{p} \times d'}$ are learnable weight matrices.

Then, we adaptively activate long-range dependency edges for each node according to the results of global structural attention scores and positional attention scores. We sample the long-range nodes using two alternative strategies: (1) Top-K and (2) Threshold control. Below, we illustrate the threshold control strategy:
\begin{equation}
    M = \text{Softmax}((\alpha U_{s}+ \beta U_{p}) \odot (\mathbf{1}-A)),
\end{equation}
Here $\odot$ is element-wise multiplication with $(\mathbf{1}- A)$ to ensure the sampling only focus on non-neighboring nodes. Then we use a differentiable sampling trick as follows:
\begin{equation}
    \hat{A} =\mathds{1}_{M>\tau}-\tilde{M}+M 
\end{equation}
where $\mathds{1}_{M>\tau}$ is an indicator function that returns 1 if \( M>\tau \). $\tilde{M}$ results from truncating the gradient of $M$. It ensures the output is one-hot and maintains the original gradients. Last, $\hat{A}$ represents an active mask used to determine which long-range dependency nodes are sampled.

Different from global attention aggregation in previous works~\cite{gt,egt,graphgps}, we employ differentiable hard sampling to restrict global message passing to key pairs of long-range nodes. This approach captures critical long-range dependencies while avoiding the overfitting and over-globalizing problems~\cite{overglobalizing} associated with the high degree of freedom in information propagation. 
Specifically, we adaptively combine the scores of global structural attention and positional attention with the attribute attention among  sampled long-range nodes: 
\begin{equation}
        z_{i,j} = \alpha U_{s}[i,j]+ \beta U_p[i,j] + \gamma \hat{U}_a[i,j],\; \forall j \in \mathcal{K}_i
\end{equation}
where $\mathcal{K}_{i}$ is the sampled long-range nodes set of node $i$, $\hat{U}_a = \text{Softmax}\left(\frac{Q_a K_a^T}{\sqrt{d}}\right)$ represents the global attention score of attribute encodings, where $Q_a$ and $K_a$ are obtained using the sampled global topology $\hat{A}$. By focusing on the sampled edges rather than all node pairs, we significantly reduce the computational complexity of computing attribute attention. After obtaining these scores, we normalize the attention scores and aggregate the global information, efficiently capturing long-range dependencies:
\begin{equation}
    \hat{H}_{i}^{\text{global}}=\sum_{j\in \mathcal{K}_{i}}\hat{z}_{ij}H_{j},\;\text{where}\;\hat{z}_{i,j} = \frac{e^{z_{i,j}}}{\sum_{k \in \mathcal{K}_{i}}e^{z_{i,k}}}\\
\end{equation}

\subsubsection{Adaptive Local-Global Integration.}
Finally, an additional adaptive module will be employed to integrate the local information and global information to obtain the output of AE $H^{(l)}$ in $l$-th layer:
\begin{equation}
    H^{(l)}= W_{\text{l}}^{(l)}\hat{H}^{\text{local}, (l)} + W_{\text{g}}^{(l)}\hat{H}^{\text{global}, (l)}
\end{equation}
where $\hat{H}^{\text{local}, (l)}, \hat{H}^{\text{global}, (l)}$ are the local and global attribute encodings in $l$-th layer, $W_{\text{l}}^{(l)}$ and $W_{\text{g}}^{(l)} \in \mathbb{R}^{d \times d}$ are learnable weight matrices in the $l$-th layer. These matrices enable us to adaptively adjust the weights for local and global information, catering to graphs with different characteristics. This adaptive weighting helps avoid local-global chaos by dynamically balancing the contributions of local and global interactions.

\subsection{Discussion}\label{sec:discuss}
\subsubsection{Complexity Analysis.}
In this subsection, we will present a theoretical complexity analysis of \our.

\paragraph{Time complexity}
For simplicity, we assume that the feature dimension remains unchanged and that the number of model layers is set to 1. 
The time complexity of \our mainly depends on four modules. The complexity of the encoder module for multi-view encodings is $\mathcal{O}(N(d^2+2K^2))$. The complexity of the local attention module is $\mathcal{O}(E(2K+d)+N(d^2+2K^2)$. The complexity of our global attention module is $\mathcal{O}(2N^{2}K + NKd+ N(d^2+2K^2))$. Finally, the complexity of multi-view attention integration and local-global integration is $\mathcal{O}(N)$ and $\mathcal{O}(Nd^{2})$ respectively. Thus the total time complexity of our method is $\mathcal{O}(2N^2K+E(d+2K)+N(4d^2+Kd+6K^2))$. Given that \( K \ll d \) and \( E \ll N^2 \), and for the sake of clarity, we omit the smaller variables. The time complexity is simplified as $\mathcal{O}(N^2K+Ed)$, which is more efficient compared to GT's $\mathcal{O}(N^{2}d)$.

\paragraph{Space complexity}
The space complexity of the encoder module for multi-view encodings is $\mathcal{O}(N(d+2K)+d^2+2K^2)$. The complexity of the local attention module is $\mathcal{O}(3E+d^2+2K^2+N(d+2K))$. The complexity of the global attention module is $\mathcal{O}(2N^{2}+E+2K^2+d^2+N(d+2K))$. Finally, the complexity of local-global integration is $\mathcal{O}(d^2)$. Thus the total space complexity is $\mathcal{O}(2N^2+4E+3d^2+6K^2+3N(d+2K))$. Given that \( K \ll d \) and \( E \ll N^2 \), and for the sake of clarity, we omit the smaller variables. The space complexity can be simplified to \(\mathcal{O}(N^2 + E + d^2 + Nd)\), which is comparable to GT's \(\mathcal{O}(N^2 + d^2 + Nd)\).

\subsubsection{Analysis for Expressivity.}
We analyze the expressive power of DeGTA and present a case study in Appendix~\ref{sec:appendidxc}. DeGTA exhibits greater expressive power than the 1-WL test through its positional and structural encodings and global attention mechanism, and the case study shows that DeGTA can produce correct results in scenarios where the coupled encoding approach yields incorrect outcomes.

\section{Experiments}
In this section, we present a comprehensive empirical investigation. Specifically, we aim to address the following research questions: 
\textbf{RQ1}: How does the proposed \our perform in node classification tasks, including both homophilic and heterophilic datasets?
\textbf{RQ2}: How does \our perform in graph-level tasks?  
\textbf{RQ3}: How effectively does \our capture long-range dependencies?
\textbf{RQ4}: How does \our enhances the interpretability of message aggregation?
\textbf{RQ5}: How does the decoupling of multi-view and local-global influence performance?
Additionally, we provide a parameter study on \( K \) and the sensitivity analysis of selection of PE/SE in Appendix~\ref{subsec:expk}. Moreover, a case study that demonstrates our enhanced expressive power and interpretability is in Figure~\ref{casestudy}.

\subsection{Experimental Setup}
\subsubsection{Datasets} We employ 10 benchmark datasets for node classification, consisting of 4 homophilic, 4 heterophilic, and 2 large-scale datasets. Additionally, we utilize 5 benchmark datasets for graph-level tasks, of which 2 are long-range graph benchmarks. Detailed information is provided in Appendix~\ref{sec:appendidxa}.

\subsubsection{Baselines}
For node classification tasks, the baselines primarily consist of three categories: (1) MPNN methods without attention such as GCN~\citep{gcn}, GCNII~\citep{gcnii}, GraphSAGE~\citep{graphsage}, and GPRGNN~\citep{gprgnn}, (2) attention-based MPNN methods including GAT~\citep{gat}, GATv2~\citep{gatv2}, ADSF~\citep{adsf}, and AERO-GNN~\citep{aerognn}, (3) Graph Transformer methods like GT~\citep{gt}, NodeFormer~\citep{nodeformer}, GraphGPS~\citep{graphgps}, NAGphormer~\citep{nagphormer}, and SGFormer~\cite{sgformer}.

For graph classification and regression tasks, we compare our method with (1) MPNN-based methods including GCN~\citep{gcn}, GIN~\citep{gin}, and GAT~\citep{gat}, (2) GT-based methods including GT~\citep{gt}, EGT~\citep{egt}, SAN~\citep{san}, GraphGPS~\citep{graphgps}, GRIT~\citep{grit}. 

\subsection{Node Classification (RQ1)}
\subsubsection{Experiments on Small and Medium-scale Datasets}
\begin{table*}[t]
\caption{Node classification performance on homophilic and heterophilic graphs. The boldface and underscore show the best and the runner-up,  respectively.}
\vspace{-1.5em}
\label{table:nodeclassif}
\begin{center}    
\renewcommand\arraystretch{1}
\resizebox{1.0\linewidth}{!}{
\begin{tabular}{lcccccccccc}
\toprule
Type      & \multicolumn{4}{c}{Homophilic graphs}                  & & \multicolumn{4}{c}{Heterophilic graphs}            \\ \cline{2-5} \cline{7-10}
Dataset       & Pubmed       & Citeseer     & Cora         &Arxiv&         & Texas       & Cornell      & Wisconsin   & Actor      & Avg   \\
\midrule
GCN~\citep{gcn}           & 79.54 ± 0.4  & 72.10 ± 0.5  & 82.15 ± 0.5  & 71.74 ± 0.3  && 65.65 ± 4.8  & 58.41 ± 3.3   & 62.02 ± 5.9  & 30.57 ± 0.7   & 65.27 \\
GCNII~\citep{gcnii}                          & 80.14 ± 0.7  & 72.80 ± 0.5  & 84.33 ± 0.5  & \underline{72.74 ± 0.2}  && 78.59 ± 6.6  & 78.84 ± 6.6  & 81.41 ± 4.7 & 35.76 ± 1.0  &73.08 \\
GraphSAGE~\citep{graphsage}                  & 78.67 ± 0.4  & 71.85 ± 0.6  & 83.76 ± 0.5  & 71.49 ± 0.3  && 82.43 ± 6.1  & 75.95 ± 5.3  & 81.18 ± 5.5 & 34.23 ± 1.0  &72.45 \\
GPRGNN~\citep{gprgnn}                        & 75.68 ± 0.4  & 71.60 ± 0.8  & 84.20 ± 0.5  & 71.86 ± 0.3  && 81.51 ± 6.1  & 80.27 ± 8.1  & 84.06 ± 5.2 & 35.58 ± 0.9  &73.10 \\
\midrule
GAT~\citep{gat}                              & 78.91 ± 0.4   & 71.89 ± 0.8  & 83.18 ± 0.5  & 71.95 ± 0.4  && 60.46 ± 6.2  & 58.22 ± 3.7  & 63.59 ± 6.1 &30.36 ± 0.9   &64.82 \\
GATv2~\citep{gatv2}                            & 79.12 ± 0.3   & 71.15 ± 1.1  & 83.88 ± 0.6  & 72.14 ± 0.5  && 60.32 ± 7.0  & 58.35 ± 3.8  & 61.94 ± 4.7 &30.27 ± 0.8   &64.65 \\
ADSF~\citep{adsf}                             & 80.21 ± 0.4   & 73.00 ± 0.4  & 84.29 ± 0.5  & 72.64 ± 0.5  && 78.15 ± 6.1  & 77.52 ± 5.9  & 69.24 ± 4.1 &34.68 ± 0.9   &71.22 \\
AERO-GNN~\citep{aerognn}                     & \underline{80.59 ± 0.5}  & \underline{73.20 ± 0.6}  & 83.90 ± 0.5  & 72.41 ± 0.4  && \underline{84.35 ± 5.2}  & 81.24 ± 6.8  & 84.80 ± 3.3 & 36.57 ± 1.1  &74.63 \\
\midrule
GT~\citep{gt}                                & 79.08 ± 0.4  & 70.16 ± 0.8  & 82.22 ± 0.6  & 70.63 ± 0.4  && 84.18 ± 5.4  & 80.16 ± 5.2  & 82.74 ± 6.0 & 34.28 ± 0.7  &72.93 \\
NodeFormer~\citep{nodeformer}                        & 79.90 ± 1.0  & 72.50 ± 1.1  & 82.20 ± 0.9  & 71.24 ± 0.6  && 81.61 ± 5.4  & \underline{82.15 ± 6.7}  & 83.17 ± 5.8 & 36.28 ± 1.2  &73.63 \\
GraphGPS~\citep{graphgps}          & 79.94 ± 0.3  & 72.75 ± 0.6  & 82.44 ± 0.6  & 70.97 ± 0.4  && 82.21 ± 6.9  & 82.06 ± 5.1  & \underline{85.36 ± 4.2} & 36.01 ± 0.9  &73.97 \\
NAGphormer~\citep{nagphormer}        & 80.57 ± 0.3  & 72.43 ± 0.8  & 84.20 ± 0.5  & 70.13 ± 0.6  && 80.12 ± 5.5  & 79.89 ± 7.1  & 82.97 ± 3.2 & 34.24 ± 0.9  &73.07 \\
SGFormer~\citep{sgformer}                            & 80.30 ± 0.6  & 72.60 ± 0.2  & \underline{84.50 ± 0.8}  & 72.63 ± 0.1  && 84.29 ± 5.2  & 81.64 ± 5.0  & 85.29 ± 5.7 & \textbf{37.90 ± 1.1}  &\underline{74.89} \\
\midrule
\our (ours)                                        & \textbf{81.19 ± 0.7}  & \textbf{73.70 ± 0.4}  & \textbf{84.79 ± 0.7}  & \textbf{73.26 ± 0.2}  && \textbf{85.44 ± 4.8}  & \textbf{83.19 ± 4.8}  & \textbf{86.95 ± 5.9} & \underline{37.87 ± 1.0} &\textbf{75.80} \\
\bottomrule
\end{tabular}
}
\vspace{-1em}
\end{center}
\end{table*}

For homophilic graphs, MPNNs benefit from their inductive bias based on the homophily assumption, resulting in higher performance compared to GTs. In contrast to MPNNs, our approach enhances the utilization of positional, structural, and attribute information in graphs by decoupling multi-view information and introducing long-range dependencies through a global sampling strategy that MPNNs cannot capture. Unlike GTs, which rely on global attention aggregation, our sampling strategy focuses on capturing only the most important and plausible long-range dependencies. Additionally, by decoupling local and global interactions and adaptively integrating them, \our places greater emphasis on local information, which is more crucial for homophilic graphs. Therefore, \our outperforms both advanced MPNNs and GTs on homophilic graphs. 


For heterophilic datasets, GTs always outperform MPNNs due to their ability to utilize global attention, which effectively filters out inter-class edges from neighboring nodes and captures disconnected yet informative nodes in the graph. Our hard sampling strategy enhances the accuracy of dependency capture, thereby mitigating the overfitting associated with global attention on small graphs. Thus \our achieves significant improvement on Texas, Cornell, and Wisconsin. For the Actor dataset, our results are on par with the SOTA model, likely because the dataset is less reliant on local information, making global attention sufficient.

In summary, \our outperforms the SOTA baselines across 7 out of 8 datasets, achieving an average absolute improvement of 1\% over the runner-up. This underscores the importance of addressing multi-view and local-global chaos.
\subsubsection{Experiments on Large-scale Datasets}
To evaluate the performance of \our on large datasets, we conducted experiments on Aminer-CS and Amazon2M. As shown in Table \ref{table:largenode}, comparing with 2 scalable GNNs and NAGphormer, \our achieves the best performance on all datasets, surpasses the runner-up over 1\% absolute improvement on Amazon2M, demonstrating its effectiveness for large-scale graphs.

This further validates the effectiveness of decoupling multi-view attention and message interaction in capturing information and learning implicit long-range dependencies at a large scale among numerous nodes. Moreover, as detailed in Appendix~\ref{subsec:expk}, a larger value of \( K \) is essential for achieving optimal performance. This is particularly crucial for large graphs, where effectively capturing long-range dependencies and maintaining a broader receptive field for both positional and structural information is necessary.

\begin{table}[t]
\caption{Node classification performance on large graphs.}
\vspace{-1em}
\label{table:largenode}
\begin{center}
\renewcommand\arraystretch{1}
\begin{tabular}{lcc}
\toprule
Dataset                              & Aminer-CS                                                   & Amazon2M                          \\
\midrule
GraphSAINT~\citep{graphsaint} & 51.91 ± 0.20                                                & 75.20 ± 0.18                      \\
GRAND+~\citep{grand}          & 54.76 ± 0.23                                                & 75.83 ± 0.21                      \\
NAGphormer~\citep{nagphormer} & 56.21 ± 0.42                                                & 77.43 ± 0.24                      \\
\midrule
\our (ours)                         & \textbf{56.38 ± 0.51}                                              & \textbf{78.49 ± 0.29}                     \\
\bottomrule
\end{tabular}
\end{center}
\vspace{-1em}
\end{table}

\subsection{Graph Classification (RQ2\&RQ3)}
\subsubsection{Experiments on Classical Datasets (RQ2)}
To evaluate the effectiveness of DeGTA across different tasks, we generalize it to both classification and regression tasks at the graph level. As shown in Table~\ref{table:graphclassif}, DeGTA consistently outperforms the baselines across all datasets, achieving significant improvements. For instance, DeGTA enhances accuracy by an absolute improvement of 4.5\% compared to the runner-up on the CIFAR10 dataset. By leveraging multi-view encodings and facilitating the adaptive interaction of local-global information, DeGTA learns suitable representations for all nodes throughout the graph, enabling mean pooling to effectively yield strong results.

\begin{table}[t]
\caption{Results for graph classification tasks.}
\vspace{-1em}
\label{table:graphclassif}
\begin{center}
\renewcommand\arraystretch{1}
\resizebox{0.95\linewidth}{!}{
\begin{tabular}{lcccc}
\toprule
\multicolumn{1}{l}{Dataset} & ZINC        && MNIST        & CIFAR10      \\
\cline{2-2} \cline{4-5}
\multicolumn{1}{l}{Metric}  & MAE↓        && \multicolumn{2}{c}{ACC↑}    \\ 
\midrule
GCN~\citep{gcn}                  & 0.367 ± 0.011 && 90.705 ± 0.218 & 55.710 ± 0.381 \\
GIN~\citep{gin}                          & 0.526 ± 0.051 && 96.485 ± 0.252 & 55.255 ± 1.527 \\
GAT~\citep{gat}                     & 0.474 ± 0.007 && 95.535 ± 0.205 & 64.223 ± 0.455 \\
\midrule
GT~\citep{gt}                & 0.226 ± 0.014 && -              & -              \\
EGT~\citep{egt}                            & 0.108 ± 0.009 && \underline{98.173 ± 0.087} & 68.702 ± 0.409 \\
SAN~\citep{san}                   & 0.139 ± 0.006 && -              & -              \\
Graphormer~\citep{graphormer}             & 0.122 ± 0.006 && -              & -              \\
GraphGPS~\citep{graphgps}                 & \underline{0.070 ± 0.004} && 98.051 ± 0.126 & \underline{72.298 ± 0.356} \\
\midrule
\our(ours)                                         & \textbf{0.059 ± 0.004} && \textbf{98.230 ± 0.112} & \textbf{76.756 ± 0.927} \\
\bottomrule
\end{tabular}
}
\end{center}
\vspace{-1em}
\end{table}

\subsubsection{Experiments on Long-range Dependencies Datasets (RQ3)}
To further evaluate \our's ability to capture long-range dependencies, we tested our method on the Long-Range Graph Benchmark (LRGB)~\citep{logb}. Specifically, we conducted experiments on two peptide graph benchmarks from LRGB, namely Peptides-func and Peptides-struct. As shown in Table~\ref{table:longrange}, GTs significantly outperform MPNNs in these datasets due to their global attention mechanism, which effectively captures long-range dependencies. However, our method more accurately captures these dependencies in the graph by decoupling local-global interactions and using an adaptive integration. This strategy shows the best performance on both datasets.
 
\begin{table}[t]
\caption{Results for graph classification tasks of long-range graph benchmarks (LRGB).}
\vspace{-1em}
\label{table:longrange}
\begin{center}
\renewcommand\arraystretch{1}
\begin{tabular}{lccc}
\toprule
\multicolumn{1}{l}{Dataset}                        &  Peptides-func        && Peptides-struct             \\
\cline{2-2} \cline{4-4}
\multicolumn{1}{l}{Metric}                         & AP↑                   && MAE↓                        \\ 
\midrule
GCN~\citep{gcn}                  &  0.5930 ± 0.0023 && 0.3496 ± 0.0013  \\
GIN~\citep{gin}                          &  0.5498 ± 0.0079 && 0.3547 ± 0.0045  \\
GAT~\citep{gat}                          &  0.5842 ± 0.0046 && 0.3504 ± 0.0028  \\
\midrule
GT~\citep{gt}                &  0.6326 ± 0.0126 && 0.2529 ± 0.0016  \\
SAN~\citep{san}                   &  0.6439 ± 0.0075 && 0.2545 ± 0.0012  \\
GraphGPS~\citep{graphgps}                 &  0.6535 ± 0.0041 && 0.2500 ± 0.0012  \\
GRIT~\citep{grit}                     &  \underline{0.6988 ± 0.0082} && \underline{0.2460 ± 0.0012}  \\
\midrule
\our(ours)                                         &  \textbf{0.7023 ± 0.0101} && \textbf{0.2437 ± 0.0014}  \\
\hline
\end{tabular}
\end{center}
\vspace{-1.5em}
\end{table}

\subsection{The Enhanced Interpretability (RQ4)}
The enhancement of multi-view interpretability provided by \our is twofold.
\begin{itemize}
    \item \emph{Node level}. As shown in the case study (Figure~\ref{casestudy}), visualizing the intermediate results of multi-view attention separately enables us to identify which information is more similar across node pairs.
    \item \emph{Graph level}. As shown in Table~\ref{table:interpretability}, analyzing the training outcomes of different attention weights allows us to discern which types of information are most critical for the specific dataset.
\end{itemize}

As shown in Table~\ref{table:interpretability}, structural and attribute weights dominate in heterophilic datasets, while positional information is minimal. This indicates that global message, attribute, and structural information are vital in heterophilic graphs, whereas local message and positional information are more beneficial in homophilic graphs. 

\begin{table}[t]
\caption{The adaptive weights of different datasets after training.}
\vspace{-1em}
\label{table:interpretability}
\begin{center}    
\renewcommand\arraystretch{1}
\resizebox{1.0\linewidth}{!}{
\begin{tabular}{lcccc}
\toprule
Dataset             & Homophily   & Positional weight    & Structural weight   & Attribute weight \\
\midrule
Pubmed       &0.66 &0.5432 &0.0763  &0.3805    \\
Citeseer     &0.63   &0.3479   &0.1022 &0.5499   \\
Cora         &0.77 &0.5759   &0.0827 &0.3414  \\
Arxiv        &0.42 &0.2893 &0.1374 &0.5733 \\
\midrule
Texas       &0.00    &0.0729    &0.1406 &0.7865 \\
Cornell     &0.02    &0.0263    &0.1663 &0.8074 \\
Wisconsin   &0.05    &0.1331    &0.2347 &0.6342 \\
Actor       &0.01   &0.0281  &0.2156  &0.7563  \\
\midrule
\end{tabular}
}
\end{center}
\vspace{-1em}
\end{table}

\subsection{Ablation Study (RQ5)}
We perform comprehensive ablation studies on the importance of designs in \our, including: (1) The decoupling of multi-view encodings. (2) The hard sampling strategy. (3) The decoupling of local-global integration.

\begin{figure}[t] 
\centering 
\includegraphics[width=0.95\linewidth]{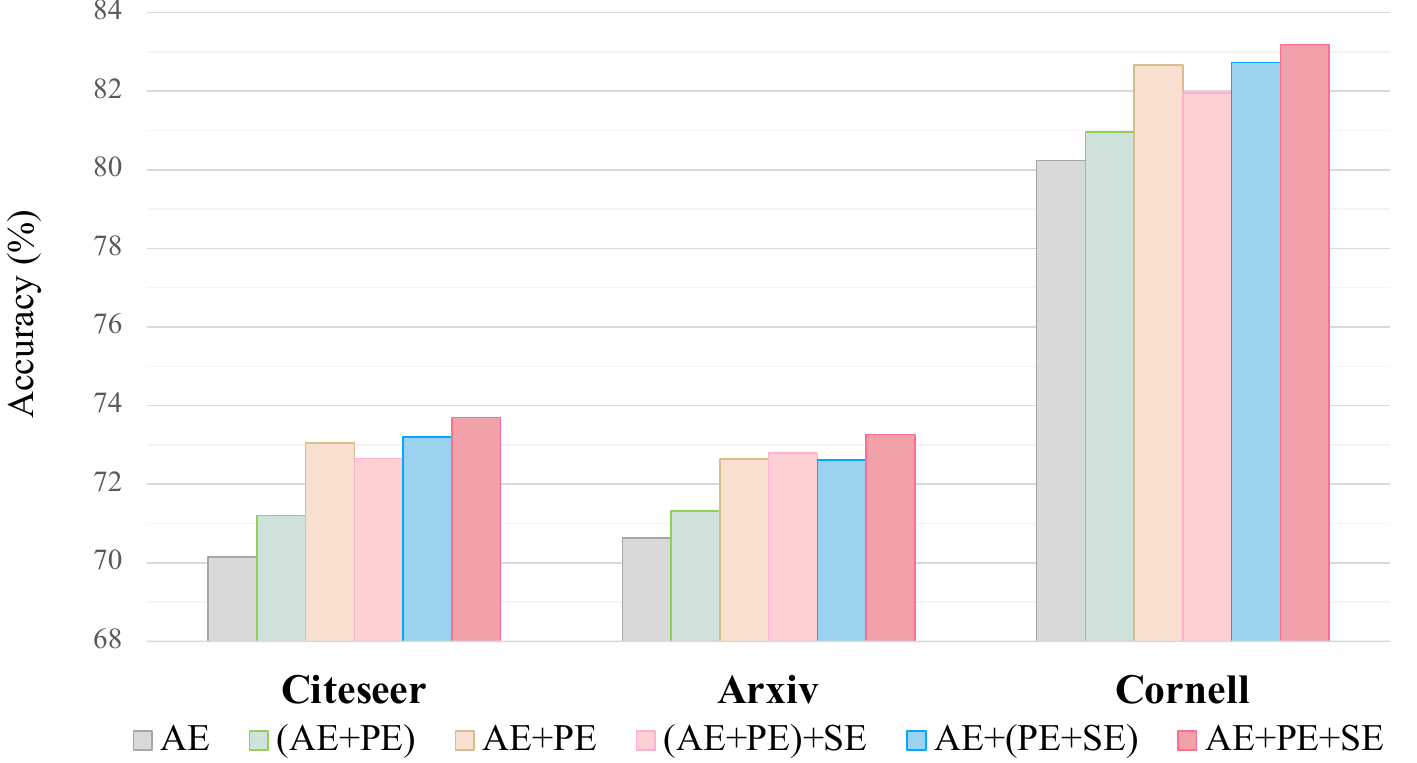}
\vspace{-1em}
\caption{The results of ablation experiments for multi-view decoupling.}
\label{fig:multiviewdecouple}
\vspace{-1.5em}
\end{figure}

\noindent\textbf{Decoupling Multi-view Encodings.}
We conduct in-depth experiments to assess the performance impact of coupling various encodings, underscoring the importance of decoupling multi-view information. We use DeGTA with only AE as the baseline. Parentheses (A+B) indicate coupled encoding, where the two encodings are concatenated and processed by a single encoder. In contrast, A+B without parentheses refers to decoupled encodings, which utilize different encoders, with attention computed separately and integrated adaptively.
For example, AE+(SE+PE) indicates that we use AE to compute attribute attention, while the fused encoding of SE and PE computes another coupled attention. As shown in Figure~\ref{fig:multiviewdecouple}, the decoupling of multi-view encoding facilitates a more comprehensive and adaptive usage of the information embedded in graphs, thus significantly improving the model's performance.

\begin{figure}[t] 
\centering 
\includegraphics[width=0.95\linewidth]{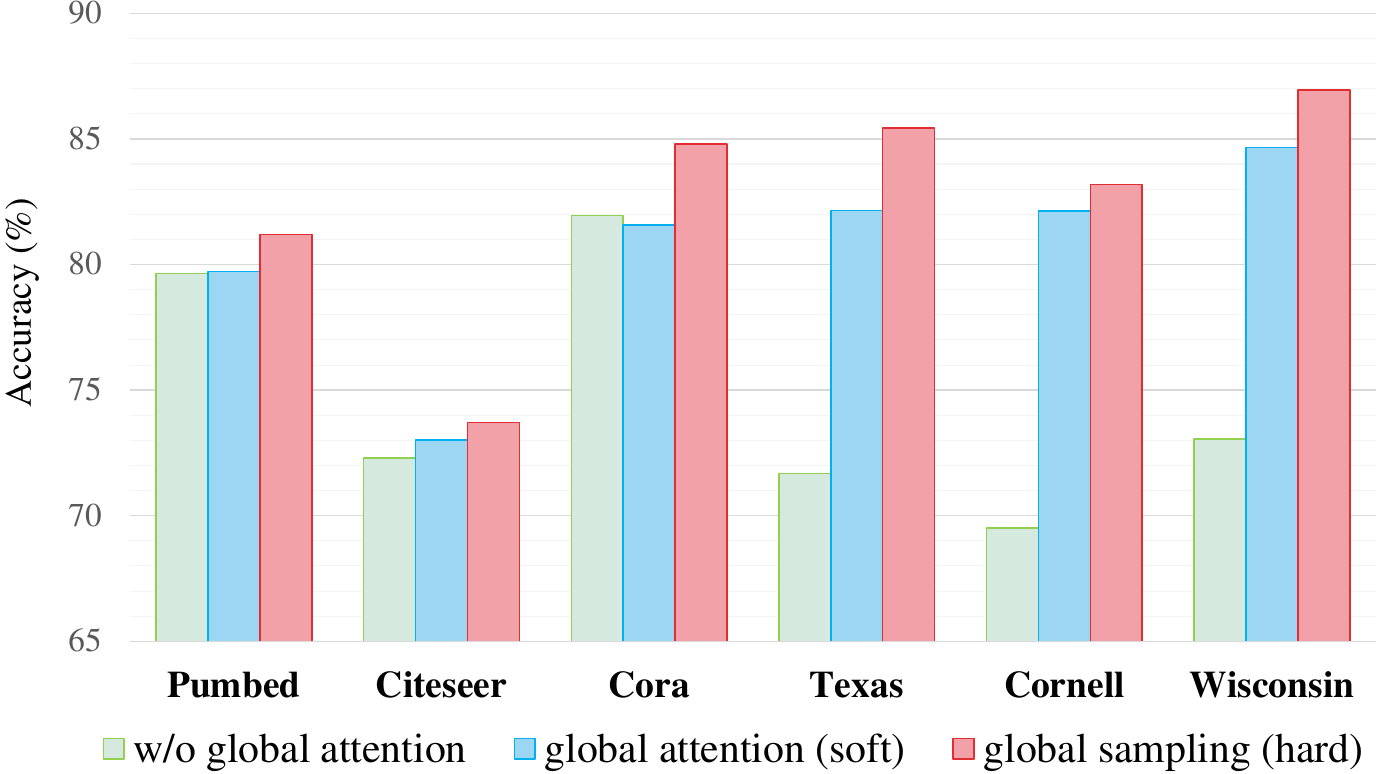}
\vspace{-1em}
\caption{The results on different strategies of global information.}
\label{exp:sampling}
\vspace{-1em}
\end{figure}

\noindent\textbf{Hard Sampling Strategy of \our.}
Unlike the global attention aggregation in Graph Transformers (GTs), we confine global message passing to key pairs of long-range nodes using a differentiable hard sampling strategy. In this section, we compare the impact of different global attention aggregation methods on DeGTA's performance. We remove the global branch of \our to serve as a baseline and evaluate the performance difference between global attention aggregation and hard-sampling aggregation. As shown in Figure~\ref{exp:sampling}, our experiments reveal that hard sampling more accurately captures long-distance dependencies and graph structures, leading to enhanced performance. Additionally, our sampling strategy reduces the time complexity associated with computing global attribute attention, as we only compute the attention for the sampled node pairs.

\noindent\textbf{Decoupling Local-global Integration.}
To evaluate the effectiveness of local-global decoupling, we conducted ablation experiments in two steps. First, we use the results of global attention instead of local attention in the local channel of DeGTA, resulting in a coupled local-global attention strategy. Second, we employed different strategies for local and global attentions but simply summed the local features and global features, thereby coupling local message passing and global attentional aggregation. As shown in Table~\ref{table:localglobal}, both changes lead to a significant decrease in performance, showing the importance of local-global decoupling and adaptive integration.

\begin{table}[t]
\caption{The ablation results for local-global decoupling. c\_attn and de\_attn denote the coupled and decoupled attention. c\_inte denotes the local-global coupling integration and ada\_inte denotes the adaptive integration.}
\vspace{-1em}
\label{table:localglobal}
\begin{center}
\renewcommand\arraystretch{1}
\resizebox{1.0\linewidth}{!}{
\begin{tabular}{lcccccccc}
\toprule
                            & Pubmed & Citeseer & Cora  & Texas & Cornell & Wisconsin \\
\midrule
c\_attn                & 79.97  & 72.40  & 82.44  & 83.59 & 80.46  & 83.27
\\
de\_attn + c\_inte     & 80.06  & 72.85  & 82.64 & 82.26 & 82.17  & 84.49    \\
de\_attn + ada\_inte    & 81.19  & 73.70  & 84.79 & 85.44 & 83.19  & 86.95    \\
\midrule
\end{tabular}
}
\vspace{-1.5em}
\end{center}
\end{table}

\section{Conclusion}
In this work, we propose a decoupled perspective to analyze attentions in graph, breaking them down into three components and two message interaction levels. This perspective helps us identify the issues of multi-view and local-global chaos in GTs. To address these challenges, we design DeGTA, a decoupled graph triple attention network. Extensive experiments demonstrate that \our achieves SOTA performance across various datasets and tasks, highlighting the effectiveness of decoupling multi-view attention and local-global interaction.

\section*{Acknowledgments}
This work was supported by the National Key Research and Development Plan of China (2023YFB4502305), and Ant Group through Ant Research Intern Program.
\bibliographystyle{ACM-Reference-Format}
\bibliography{refs_sim}


\appendix
\begin{table}[h]
\caption{Details of node classification datasets.}
\label{nodebenchmark}
\begin{center}    
\vspace{-1em}
\renewcommand\arraystretch{1}
\resizebox{1.0\linewidth}{!}{
\begin{tabular}{lcccccc}
\hline
Dataset             & \#Nodes         & \#Edges    & \#Features   &\#Classes &Split(\%)\\
\hline
Pubmed~\cite{citionnetworks}       &19,717 &44,324 &500  &3 &0.3/2.5/5.0\\
Citeseer~\cite{citionnetworks}     &3,327   &4,552   &3,703 &6 &5.2/18/37\\
Cora~\cite{citionnetworks}         &2,708   &5,728   &1,433 &7 &3.6/15/30\\
Arxiv~\cite{ogb}        &169,343&1,166,243 &128 &40 &53.7/17.6/28.7\\
\hline
Texas~\cite{wikicsactor}       &183    &279    &1,703 &5 &48/32/20\\
Cornell~\cite{wikicsactor}     &183    &277    &1,703 &5 &48/32/20\\
Wisconsin ~\cite{wikicsactor}  &251    &450    &1,703 &5 &48/32/20\\
Actor~\citep{wikicsactor}      &7,600   &26,659  &932  &5 &48/32/20\\
\hline
AMiner-CS~\citep{aminercs}   &593,486   &6,217,004  &100 &18 &20/30/50~\citep{grand}\\
Amazon2M~\citep{amazon2m}    &2,449,029 &61,859,140 &100 &47 &20/30/50~\citep{grand}\\
\hline
\end{tabular}
}
\vspace{-1em}
\end{center}
\end{table}

\section{More Experiment details}
\label{sec:appendidxa}

\begin{table*}[t]
\caption{Details of graph classification datasets.}
\label{graphbenchmark}
\begin{center}    
\vspace{-1em}
\renewcommand\arraystretch{1}
\begin{tabular}{lccccccc}
\hline
Dataset             & \#Graphs   & Avg.\#Nodes     & Avg.\#Edges    & Prediction task  &\#Classes &Metric &Split(\%)\\
\hline
ZINC~\cite{graphtask}            &12,000  &23   &25  &regression &-  &Mean Abs. Error &83.3/8.3/8.3\\
MINST~\cite{graphtask}           &70,000  &71   &565 &classif.   &10 &Accuracy        &78.6/7.1/14.2\\
CIFAR10~\cite{graphtask}         &60,000  &118  &941 &classif.   &10 &Accuracy        &75/8.3/16.7 \\
\hline
Peptides-func~\citep{logb}   &15,535  &151 &307 &classif.    &10 &Avg. Precision  &70/15/15 \\
Peptides-struct~\citep{logb} &15,535  &151 &307 &regression  &11 &Mean Abs. Error &70/15/15\\
\hline
\end{tabular}
\vspace{-1em}
\end{center}
\end{table*}

\subsection{Details of Datasets}
We employs 10 node classification benchmark datasets, among which 4 are homophilic, 4 are heterophilic, and 2 are large-scale datasets. on the other hand, we employs 5 graph classification benchmark datasets, among which 2 are long range graph benchmark. The train-validation-test splits utilized are those which are publicly accessible. Details of these benchmark are shown in Table~\ref{nodebenchmark} and Table~\ref{graphbenchmark}.

\subsection{Details of hyperparameter}
Experimental results are reported on the hyperparameter settings in our code link, where we choose the settings that achieve the highest performance on the validation set. We choose hyperparameter grids that do not necessarily give optimal performance, but hopefully cover enough regimes so that each model is reasonably evaluated on each dataset. 

\begin{figure}[t] 
\centering 
\includegraphics[width=1\linewidth]{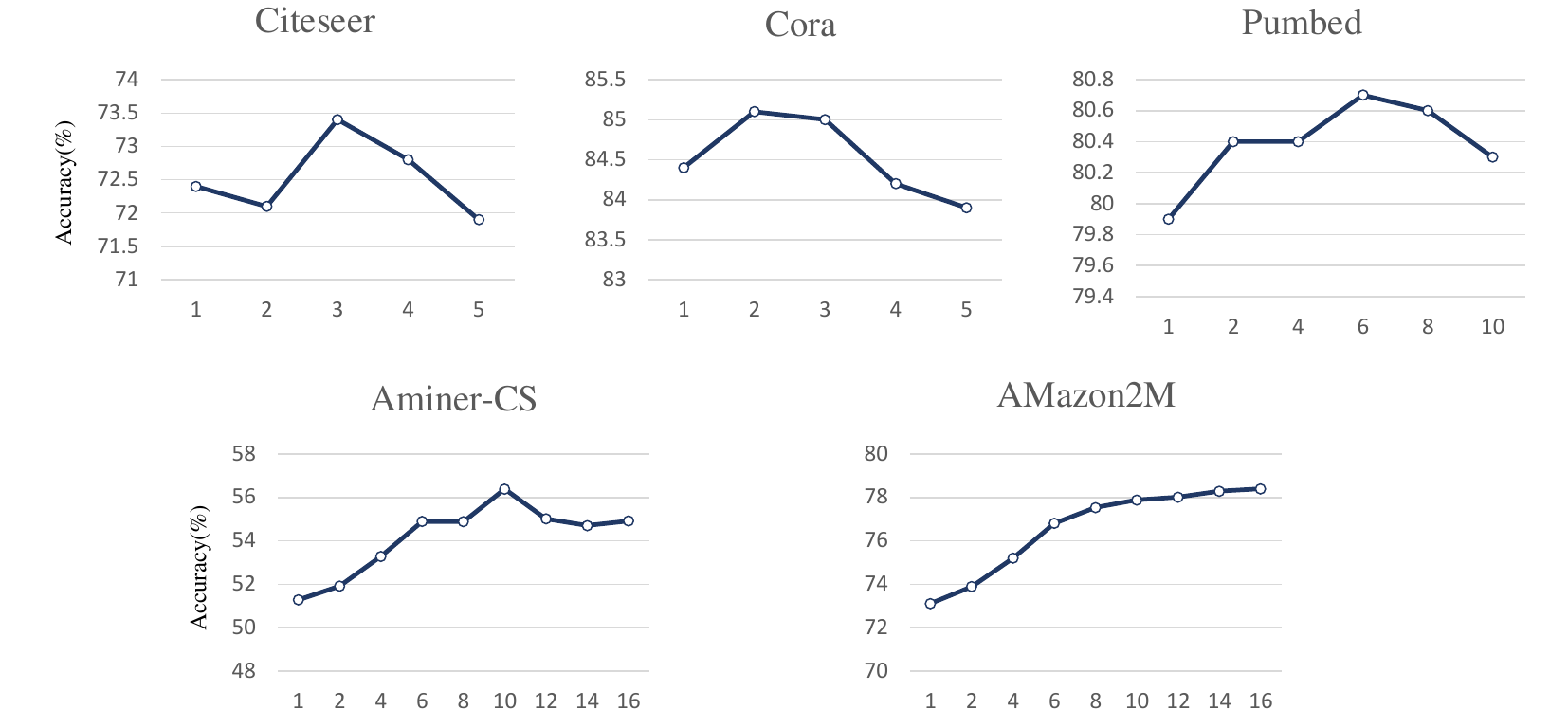} 
\caption{The results of experiments for the hyperparameter K}
\label{expofK} 
\vspace{-1.5em}
\end{figure}

\textbf{Parameter study on $K$.}
\label{subsec:expk}
We conducted experiments on 3 small datasets and 2 large datasets for the hyperparameter $K$, \eg, the dim of initial SE, PE, and the numeber of sampled long-range nodes of each node. As shown in Figure~\ref{expofK}, we can observe that the values of $K$ are different for each dataset to achieve the optimal performance, since different graph exhibit different importance for positional information, structural information, and long-range dependency (global information). 

For smaller datasets such as Citeseer, Cora, and Pubmed, optimal performance is achieved when \( K \) takes on a small value, as a small \( K \) is sufficient to capture the necessary field of view for positional and structural information. However, as \( K \) increases up to the graph diameter, nodes gain a perspective of nearly the entire graph, which results in a sharp drop in performance akin to over-smoothing. On the other hand, in homophilic graphs, the emphasis is on local information, thus a small \( K \) is enough to capture sufficient long-distance dependencies.

For large datasets \eg Aminer-CS and AMazon2M, larger \( K \) is necessary to achieve optimal performance. This is because it is crucial for large graphs to capture long-range dependencies and maintain a broader receptive field for positional and structural information.

\textbf{Sensitivity analysis of the selection of PE/SE.}
\label{subsec:exppe/se}
DeGTA is robust and insensitive to the selection of PE/SE and the multi-view encoder. We implemented a simple combination of Jaccard/RWSE (well-established in previous works), utilizing MLP as the encoder for various inputs, yet achieved substantial performance.

Table~\ref{exp:pe/se} demonstrating that arbitrary combinations of encodings can surpass the performance of coupled approaches within the \our framework. Thus we not selectively choose from existing methods to find a combination that perform well, but offer a guiding decoupled framework which is robust to all settings.

\begin{table}[t]
\caption{Ablation experiment of different PE/SE in Arxiv. DeGTA is insensitive to the choice of encoding for multi-view information, but sensitive to decoupling.}
\label{exp:pe/se}
\begin{center}    
\vspace{-1em}
\renewcommand\arraystretch{1}
\resizebox{1.0\linewidth}{!}{
\begin{tabular}{ccccc}
\toprule
PE             & SE   &DeGTA (decoupling)     & coupling (PE+SE)    & coupling (AE+PE)  \\
\midrule
RWPE     &RWSE  &73.31   &71.69  &71.38 \\
RWPE     &DSE  &72.97   &70.99   &70.94   \\
Jaccard  &RWSE &73.26  &72.13 &72.02   \\
Jaccard  &DSE  &73.12	 &72.08 &70.82  \\
LapPE    &RWSE  &73.09	 &71.24 &71.36 \\
LapPE    &DSE  &72.54 &70.98 &69.97  \\
\midrule
\end{tabular}
}
\vspace{-1.5em}
\end{center}
\end{table}

\section{Details of Initialization Strategies of PE/SE}
\label{sec:appendidxb}
In this section, we provide a detailed description of the encoding initialization strategies discussed in Section~\ref{subsec:attn}.

\paragraph{1) Strategies for Structural Encodings.} As discussed in Section~\ref{subsec:attn}, structural encodings should capture the ability to perceive nodes' surrounding structure. In our method, we provide several strategies to achieve this objective: (1) Random-Walk Structural Encoding (RWSE), (2) Node Degree Encoding (DSE), and (3) Topology Counting Encoding (TCSE).

Specifically, RWSE is computed by the probability of arriving at the node itself with $0$ step to $K-1$ step wandering. It reflect pure structure information of topology information of the K-hop receptive field.
\begin{equation}
\label{eq1}
s_{i}^{RWSE}=[I,\hat{A},\hat{A}^{2},\cdots\hat{A}^{K-1}]_{i,i}\in \mathbb{R}^{K}
\end{equation}

DSE is computed by the indegree and outdegree of the node, while TCSE is computed by counting the topological structure of its K-hop subgraph, such as the count of triangles, quads, and rings.

\paragraph{2) Strategies for Positional Encodings.} To introduce the ability for a node to perceive its position relative to other specific nodes, we provide several off-the-shelf strategies that can reflect a node's distance or intersection relationships with other specific nodes: (1) Laplacian Positional Encoding (LapPE), (2) Relative Random Walk Probabilities (RWPE), and (3) Jaccard Encoding (JaccardPE).

Specifically, LapPE is a general method to encode node positions. For each node, its Laplacian PE is the $K$ smallest non-trivial eigenvectors.
\begin{equation}
p_{i}^{LapPE} = [\phi_{0,i},\phi_{1,i},\phi_{2,i},\cdots\phi_{K-1,i}]\in \mathbb{R}^{K}
\end{equation}
where $\phi_{m,i}$ is the $i$-th row of normalized eigenvector associated to the $m$-th lowest eigenvalue $\lambda_{m}$.

RWPE is computed by the probability of one node arriving other nodes within $K$ steps. It reflect the positional interaction of a node with other nodes in the graph.
\begin{equation}
\label{eq2}
\begin{aligned}
\breve{p}_{i}^{RWPE} &= [I,\hat{A},\hat{A}^{2},\cdots\hat{A}^{K-1}]_{i}\in \mathbb{R}^{K\times N}\\
p_{i}^{RWPE} &= \mathcal{F}(\breve{p}_{i}^{RWPE})
\end{aligned}
\end{equation}
where $\mathcal{F}(\cdot): \mathbb{R}^{K \times N} \mapsto \mathbb{R}^{K} $ is a encoder to condense information.

Jaccard coefficient is used to compare similarities between different sets. It is the ratio of the size of the intersection of A and B to the size of the concatenation of A and B. In graph domain, we can use Jaccard coefficient as positional similarity between global node pairs. As follows:
\begin{equation}
J(p_{i},p_{j})=\frac{{\textstyle\sum_{k\in(V_{i}\cup V_{j})}\min(p_{ik},p_{jk}) } }{\sum_{k\in(V_{i}\cup V_{j})}\max (p_{ik},p_{jk}) }
\end{equation}
Similarly, we would like to represent our positional attention in terms of intersection, concatenation, and importance relationships between nodes, thus we use the shortest distance encoding as JaccardPE, and adopt the weight of node pairs with a Gaussian decay, i.e., $p_{i,j}^{Jaccard}=exp(-\frac{distance(i,j)^{2}}{2h^{2}})$. The algorithm is as follows: 

\begin{algorithm}
\vspace{-0.3em}
	\renewcommand{\algorithmicrequire}{\textbf{Input:}}
	\renewcommand{\algorithmicensure}{\textbf{Output:}}
	\caption{Fast Jaccard encoding computing}
	\label{alg1}
	\begin{algorithmic}[1]
        \REQUIRE Normalized adjacency matrix $\tilde{A}$; Receptive field $K$
        \ENSURE  Initialized positional encoding $P^{Jaccard}$
		\STATE Initialization:$P^{Jaccard} \leftarrow 0, B = I$
		\FOR{$k=0$ to $K-1$}
		\FOR{$i,j$ in range($N$)}
		\IF {$B_{i,j}>0$ and $P_{i,j}^{Jaccard}=0$}
		\STATE            $P_{i,j}^{Jaccard}=exp(-\frac{k^{2}}{2})$
        \ENDIF
		\ENDFOR
        \STATE $B=B\tilde{A}$
		\ENDFOR 
		\RETURN $P^{Jaccard}$
	\end{algorithmic}  
\end{algorithm}

\section{Expressive power of DeGTA}
\label{sec:appendidxc}
\textbf{1-Weisfeiler-Leman (1-WL) Test and MPNNs.} The 1-WL test is a node-coloring algorithm within the hierarchy of WL heuristics used for graph isomorphism. The limitations of MPNNs in distinguishing non-isomorphic graphs were rigorously analyzed in the work of \citep{gin}. This analysis highlights the established equivalence between MPNNs and the 1-WL isomorphism test, as detailed by \citep{wltest}. As a result, MPNNs may perform poorly on graphs that exhibit multiple symmetries within their original structure, including node and edge isomorphisms.

\textbf{Expressive power of DeGTA.}
In the DeGTA architecture, we employ decoupled structural and positional encodings, both of which offer greater expressive power than the 1-WL test through the global positional and structural attention mechanism, as demonstrated in \citep{san,graphgps}. Thus, DeGTA demonstrates excellent performance on graphs that exhibit multiple symmetries within their original structure, including node and edge isomorphisms. It is capable of distinguishing any pair of non-isomorphic graphs as well as CSL graphs, which cannot be learned by the 1-WL test or MPNNs. Further, compared to the GT with SE/PE, through the decoupling of PE, SE and AE, DeGTA can distinguish some graphs with more sensitive positional, structural and attribute information that coupled PS/SE cannot learn, as shown in case study. 

\begin{figure}[t] 
\vspace{-0.5em}
\centering 
\includegraphics[width=1\linewidth]{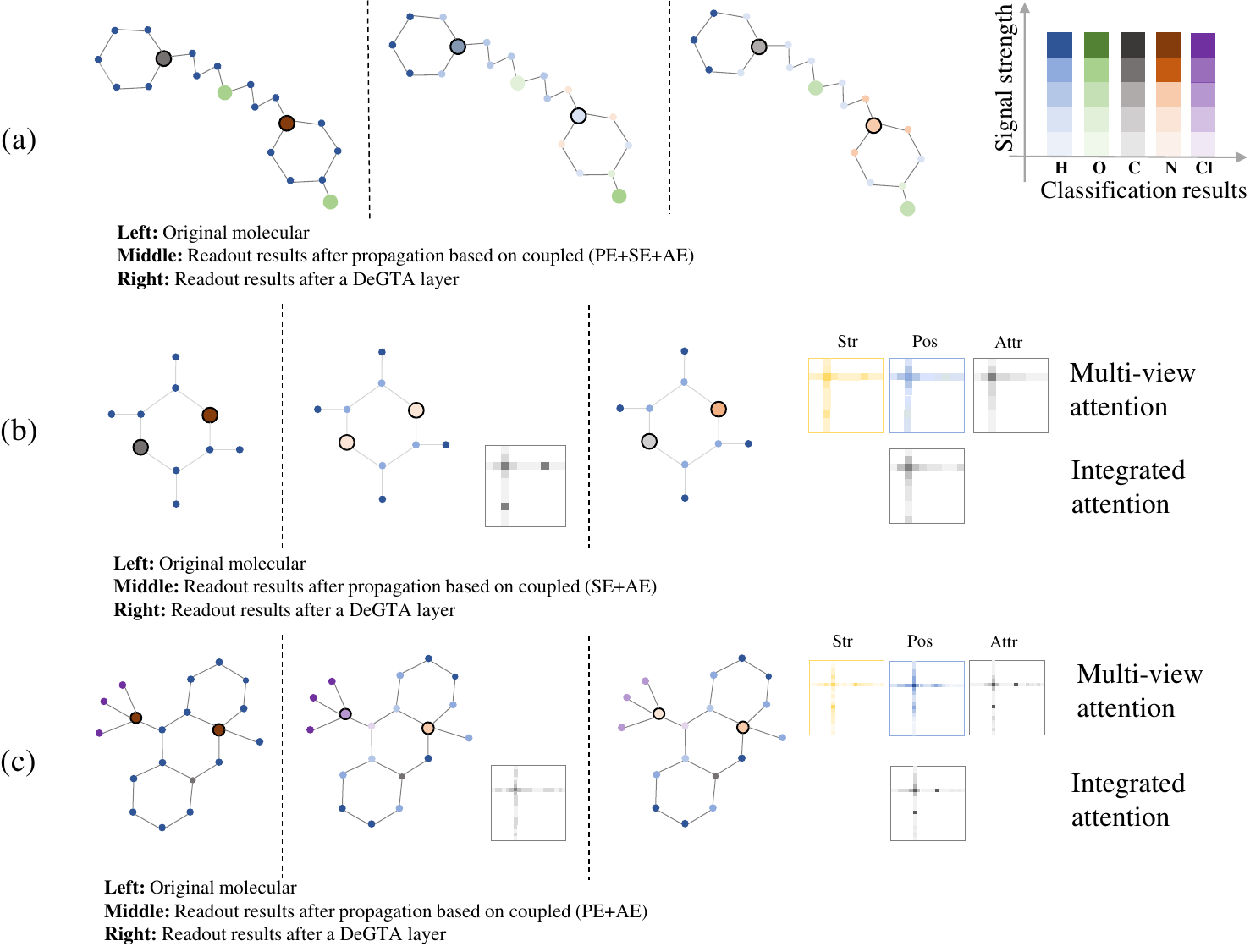} 
\caption{Our case study of molecular classification. For each case (left), we focus on pairs of nodes highlighted by a black border. These node pairs will produce incorrect classification results when using coupled encoding for global attention (middle), whereas DeGTA can produce correct results through decoupled multi-view attention and adaptive aggregation (right).}
\label{casestudy} 
\vspace{-1em}
\end{figure}

\textbf{Case study.} As shown in Figure~\ref{casestudy}, we present additional visualization examples of the molecular graph propagation process to demonstrate the necessity of decoupling positional, structural, and attribute encodings. In these experimental cases, the DeGTA layer often produces correct results in scenarios where the coupled encoding approach yields incorrect outcomes.

For instance, in case (b), we observe that the coupling of SE and AE results in high attention between the highlight pair, leading to erroneous classification outcomes (middle). However, with multi-view decoupling, we see that the pair exhibit high structural attention but low attribute and positional attention. This allows adaptive integration to learn a low overall attention between the two nodes, effectively suppressing message passing and yielding the correct classification (right). In case (c), the coupling of PE and AE results in low attention between the pair, again leading to incorrect classification (middle). In contrast, the decoupled attention shows that the node pairs possess high structural and attribute attention but low positional attention. The high structural attention activates long-range sampling between the pair, and the adaptive integration demonstrates a high importance to attribute information, resulting in a high overall attention between the two nodes and achieving the correct classification (right).

\end{document}